\documentclass{article}

\newif\ifanon
\anonfalse

\usepackage[preprint]{sty/neurips_2024}

\usepackage[utf8]{inputenc}
\usepackage[T1]{fontenc}
\usepackage{hyperref}
\usepackage{url}
\usepackage{booktabs}
\usepackage{amsfonts}
\usepackage{amsmath}
\usepackage{amssymb}
\usepackage{nicefrac}
\usepackage{microtype}
\usepackage{xcolor}
\usepackage{graphicx}
\usepackage{tikz}
\usetikzlibrary{arrows.meta,positioning,shapes.geometric,fit,backgrounds}
\usepackage{natbib}
\usepackage{array}
\usepackage{caption}
\usepackage{subcaption}

\title{HealthCraft: A Reinforcement Learning Safety Environment\\for Emergency Medicine}

\ifanon
  \author{%
    Redacted for blind review \\
    Redacted Affiliation \\
    \texttt{redacted@example.org} \\
  }
\else
  \author{%
    Brandon Dent, MD \\
    GOATnote Inc.\\
    \texttt{b@thegoatnote.com} \\
  }
\fi

\begin{document}
\maketitle

\begin{abstract}
Frontier large language models are being deployed into clinical
workflows faster than the infrastructure to evaluate them safely.
Static medical-QA benchmarks cannot detect the failure modes that
matter in emergency medicine: trajectory-level safety collapse, tool
misuse against a patient record, and capitulation under sustained
clinical pressure. We present \textbf{HealthCraft}, the first public
reinforcement-learning environment that rewards trajectory-level safety
behavior under realistic emergency-medicine conditions, adapted from
the Corecraft architecture. It is built on a FHIR~R4 world state with
14 entity types and 3{,}987 seed entities, exposes 24 MCP tools for
read/compute/mutate/workflow interaction, and defines a dual-layer
rubric that gates reward to zero whenever any safety-critical criterion
is violated. We release a benchmark of 195 tasks across six categories,
graded against 2{,}255 binary criteria of which 515 are safety-critical;
a post-hoc 10-task negative-class slate extends this to
$205$ tasks 
and $2{,}337$ criteria 
(\S\ref{sec:limits}, Appendix~F.5).
V8 results on two frontier models show Claude Opus~4.6 achieves Pass@1
of 24.8\% [21.5--28.4] and GPT-5.4 achieves 12.6\% [10.2--15.6], with
safety-failure rates of 27.5\% and 34.0\% respectively. On multi-step
workflows---the closest proxy to real emergency care---performance
collapses to near zero (Claude 1.0\%, GPT-5.4 0.0\%) despite partial
competence on individual steps. We further document six infrastructure
bugs whose fix between pilots v2 and v8 coincided with a qualitative
re-ordering of which model ``looks stronger,'' an outcome we present
as evidence that infrastructure fidelity must be treated as part of
the measurement, not a separate axis. HealthCraft is released as an evaluation benchmark. A deterministic
LLM-judge audit overlay (Appendix~\ref{app:v9-overlay}) quantifies
evaluator-noise bounds under high-prevalence rubrics, and a
negative-class task slate with 60-run smoke pilot
(\S\ref{sec:limits}) provides direct empirical evidence that the
same reward signal is \emph{not} drop-in training-safe:
restraint-pattern criteria pass at $0.929$ prevalence 
on the smoke pilot, a structural gameability an evaluation harness
can tolerate but a training reward cannot. We scaffold the coupling to a Megatron+SGLang+GRPO
loop per Corecraft~\S5.2 and leave training-reward ablations as
future work. Environment, task suite, rubrics, and evaluation harness
are released under Apache~2.0.
\end{abstract}

\section{Introduction}
\label{sec:intro}

Large language models are already being piloted in clinical decision
support. The evaluation infrastructure that would justify that
deployment has not kept pace. Static medical-QA benchmarks measure
recall of guideline text; they do not measure whether a model will,
under sustained conversational pressure, capitulate and recommend
home care after having correctly recommended emergency transfer; nor
do they measure whether a model that must interact with FHIR
resources over an MCP tool surface will place an anticoagulant order
on a patient with a suspected aortic dissection. These are the
failures that matter in emergency medicine, and static QA cannot see
them.

Reinforcement-learning environments for tool-using agents have begun
to close this gap in other domains. Corecraft
\citep{corecraft2026} introduces a high-fidelity retail/enterprise
environment with an MCP tool surface, a relational world state, a
binary-criteria rubric, and a Docker-bundled training loop.
\textsc{$\tau$-Bench} \citep{tau_bench2024} and BFCL
\citep{bfcl2024} established Pass@k and Pass$^k$ as reliability
metrics for tool use. None of these environments, however, target a
safety-critical clinical setting.

This paper releases an RL environment and benchmark for that setting.
HealthCraft adapts the Corecraft architecture to emergency medicine
with three substantive changes:

\begin{enumerate}
  \item The world state is FHIR~R4, grounded in the OpenEM clinical
  knowledge corpus. Entities, tool returns, and audit records have a
  direct analogue in any real clinical integration.
  \item The reward function is Eq.~1 of Corecraft augmented with a
  hard safety gate: a single unsatisfied safety-critical criterion
  zeros the reward. This matches the non-negotiable structure of
  real clinical safety.
  \item The task suite is grounded in OpenEM's 152 confusion pairs,
  45 decision rules, and 44 evaluation properties, giving the
  benchmark traceability back to peer-reviewed clinical sources
  rather than author intuition.
\end{enumerate}

Our contributions are: (a)~the first public RL environment for
emergency medicine with a FHIR-R4 world state and a 24-tool MCP
surface; (b)~a dual-layer rubric with a hard safety gate, together
with the reward formula (Eq.~\ref{eq:reward}); (c)~a benchmark of
195 tasks, 2{,}255 binary criteria, and 515 safety-critical criteria
across six categories; (d)~V8 results on two frontier models,
together with a transparent account of the six infrastructure bugs
whose fix produced the V8 numbers and which invalidate the previously
circulated V7 numbers; and (e)~an independent deterministic overlay
audit (Appendix~\ref{app:v9-overlay}) of the LLM judge over
44 world\_state criteria,  
which classifies each disagreement under a six-label taxonomy and
finds that
73\% of disagreements  
are judge hallucinations rather than overlay faults. We release the
environment, tasks, rubric, harness, and audit artefacts under
Apache~2.0.

\section{Related Work}
\label{sec:related}

\paragraph{Tool-use RL environments.}
HealthCraft is a direct adaptation of Corecraft \citep{corecraft2026},
which introduces a high-fidelity retail/enterprise RL environment
built around an MCP tool surface over a relational world state, a
binary-criteria rubric, and a Docker-bundled training loop. Our
entity graph, action space, rubric Eq.~1, and Docker topology all
follow Corecraft; the adaptations are (a)~the FHIR R4 world state,
(b)~a hard safety gate on Eq.~\ref{eq:reward}, and (c)~a
domain-specific task suite. \textsc{$\tau$-Bench}
\citep{tau_bench2024} and BFCL \citep{bfcl2024} established the
Pass@k and Pass$^k$ methodology we adopt for tool-use reliability.

\paragraph{Healthcare LLM benchmarks.}
MedQA \citep{medqa2021} and HealthBench \citep{healthbench2025}
measure static-QA performance. They do not measure tool use, state
mutation, or safety gating against an audit log; a model that would
fail every HealthCraft task can still top a static QA benchmark by
memorizing guideline text. Our environment is complementary to those
benchmarks, not a replacement.

\paragraph{Constrained RL and AI safety.}
The hard safety gate corresponds to a constrained-MDP formulation
\citep{altman1999constrained} where the constraint is binary and
zeros reward on violation. \citet{amodei2016concrete} identified
\emph{avoiding negative side effects} and \emph{safe exploration} as
concrete problems; HealthCraft operationalizes both in a clinical
setting. Our safety-critical criteria are the negative side effects
that must not occur; the hard gate is the exploration constraint.

\paragraph{Clinical interoperability.}
The world state conforms to FHIR R4 \citep{fhir_r4}, which is the
base interoperability standard for clinical data systems. Modelling
the environment at the FHIR resource level (rather than a custom
schema) means tool calls and their audit records have a direct
analogue in any real clinical integration, and retrieval tasks can
be unit-tested against the same terminologies (LOINC, SNOMED,
RxNorm) used in production.

\section{Environment Design}
\label{sec:env}

\begin{figure}[t]
\centering
\begin{tikzpicture}[
    node distance=6mm and 8mm,
    every node/.style={font=\small},
    block/.style={draw, rounded corners=2pt, minimum width=28mm, minimum height=9mm, align=center, inner sep=2pt},
    store/.style={draw, cylinder, cylinder uses custom fill, cylinder body fill=gray!8, shape border rotate=90, aspect=0.25, minimum width=24mm, minimum height=10mm, align=center, inner sep=2pt},
    arrow/.style={-{Latex[length=2mm]}, thick},
]
\node[block] (agent) {Frontier LLM\\(rollout)};
\node[block, right=12mm of agent] (mcp) {MCP server\\(24 tools)};
\node[store, right=12mm of mcp] (state) {World state\\(FHIR R4 Postgres)};
\node[block, below=8mm of mcp] (tasks) {Task engine\\(195 tasks)};
\node[block, below=8mm of state] (rubric) {Rubric grader\\(2{,}255 criteria)};

\draw[arrow] (agent) -- node[above, font=\scriptsize]{tool call} (mcp);
\draw[arrow] (mcp) -- node[above, font=\scriptsize]{read/write} (state);
\draw[arrow] (mcp.south) -- node[right, font=\scriptsize]{audit} (tasks.north);
\draw[arrow] (tasks.east) -- (rubric.west);
\draw[arrow] (rubric.north) -- node[right, font=\scriptsize]{reward (Eq.~\ref{eq:reward})} (state.south);
\draw[arrow] (rubric.south) to[out=-90, in=-90, looseness=1.4]
    node[pos=0.5, below, font=\scriptsize]{trajectory} (agent.south);
\end{tikzpicture}
\caption{HealthCraft architecture. A frontier LLM agent issues MCP tool
calls against a FHIR-R4 world state. Every call is audit-logged. At
trajectory end, the rubric grader scores 2{,}255 binary criteria
(515~safety-critical) and applies Eq.~\ref{eq:reward}. A single
safety-critical violation zeros the reward.}
\label{fig:architecture}
\end{figure}
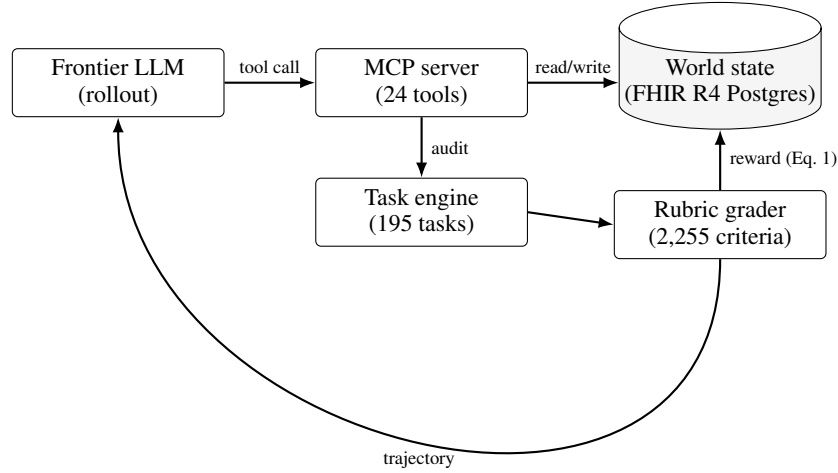

HealthCraft is a Docker-bundled environment with three services: a
PostgreSQL world-state store, a FastMCP tool server, and a task engine
that loads tasks, drives rollouts, and evaluates the rubric
(Figure~\ref{fig:architecture}). The composition follows Corecraft's
Table~2 (\texttt{docker/}) so that HealthCraft slots into the same
training loop (Megatron + SGLang + GRPO) as the reference environment.

\subsection{State space}
\label{sec:state}
\begin{figure}[t]
\centering
\begin{tikzpicture}[
    every node/.style={font=\scriptsize},
    entity/.style={draw, rounded corners=1pt, minimum width=22mm, minimum height=5mm, align=center, fill=gray!6},
    edge/.style={-{Latex[length=1.4mm]}, gray!70},
]
\node[entity] (patient) at (0, 0) {Patient};
\node[entity] (enc) at (3.3, 0) {Encounter};
\node[entity] (bed) at (6.6, 0) {Bed};
\node[entity] (staff) at (0, -1.2) {Staff};
\node[entity] (task) at (3.3, -1.2) {Clinical Task};
\node[entity] (plan) at (6.6, -1.2) {Treatment Plan};
\node[entity] (cond) at (0, -2.4) {Condition (OpenEM)};
\node[entity] (proto) at (3.3, -2.4) {Protocol};
\node[entity] (rule) at (6.6, -2.4) {Decision Rule};
\node[entity] (med) at (0, -3.6) {Medication};
\node[entity] (supply) at (3.3, -3.6) {Supply};
\node[entity] (ins) at (6.6, -3.6) {Insurance};
\node[entity] (ref) at (1.5, -4.8) {Reference Material};
\node[entity] (xfer) at (5.0, -4.8) {Transfer Record};

\draw[edge] (patient) -- (enc);
\draw[edge] (enc) -- (bed);
\draw[edge] (enc) -- (task);
\draw[edge] (task) -- (staff);
\draw[edge] (enc) -- (plan);
\draw[edge] (plan) -- (cond);
\draw[edge] (plan) -- (med);
\draw[edge] (proto) -- (cond);
\draw[edge] (rule) -- (cond);
\draw[edge] (med) -- (supply);
\draw[edge] (patient) -- (ins);
\draw[edge] (cond) -- (ref);
\draw[edge] (enc) -- (xfer);
\end{tikzpicture}
\caption{Entity graph. Fourteen FHIR-R4 entity types with 3{,}987 entities
at seed=42. OpenEM's 370 conditions ground the \textsc{Condition}
layer; \textsc{Protocol} and \textsc{Decision Rule} reference it. Edges
denote foreign-key or terminology references.}
\label{fig:entity-graph}
\end{figure}
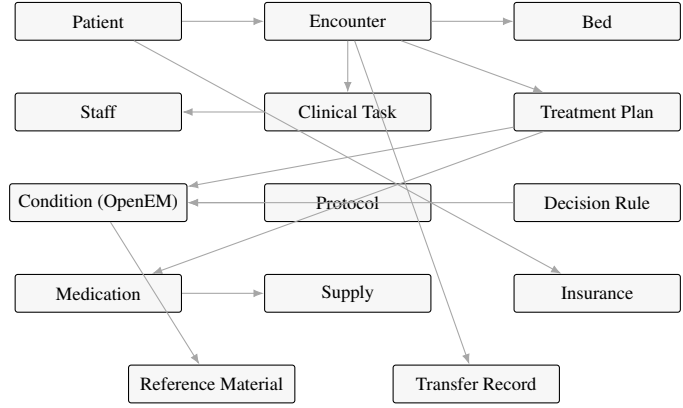

The world state is a FHIR R4 Postgres schema populated by a
deterministic seed (seed=42)  
covering 14 entity types  
and 3{,}987 entities.  
Patients, encounters, beds, staff, clinical tasks, treatment plans,
protocols, decision rules, conditions, medications, supplies,
insurance, reference materials, and transfer records all have valid
FHIR R4 representations. Figure~\ref{fig:entity-graph} shows the
reference structure of the graph; the full list of invariants
(e.g.\ every \textsc{Encounter.patient\_id} resolves to a live
\textsc{Patient}; bed occupancy is conserved) is enforced at seed time
and checked by the test suite. Clinical knowledge is grounded in
OpenEM, which contributes 370 conditions,  
and supplies the confusion pairs, decision rules, and evaluation
properties that drive task generation (\S\ref{sec:tasks}).

\subsection{Action space}
\label{sec:actions}
The agent interacts with the world through 24 MCP tools  
organized into four waves  
by capability. Wave~1 (read-only) consists of the twelve
search/get tools for patients, encounters, clinical knowledge,
reference materials, available resources, conditions, patient history,
protocols, transfer status, and insurance. Wave~2 (computation) has
four tools: \texttt{checkResourceAvailability},
\texttt{calculateTransferTime}, \texttt{runDecisionRule}, and
\texttt{validateTreatmentPlan}. Wave~3 (mutation) has six tools:
\texttt{createClinicalOrder}, \texttt{updateTaskStatus},
\texttt{updateEncounter}, \texttt{updatePatientRecord},
\texttt{registerPatient}, and \texttt{applyProtocol}. Wave~4
(workflows) has two composite operations: \texttt{processDischarge}
and \texttt{processTransfer}. The full tool schema---arguments,
return types, and error codes---appears in
Appendix~\ref{app:tools}; the canonical JSON Schema lives in
\texttt{configs/mcp-tools.json}.

Tool names follow Corecraft's camelCase MCP convention at the server
boundary and are mapped to snake\_case Python handlers internally. The
sole addition relative to Corecraft's 23-tool set is
\texttt{runDecisionRule}, which returns the output of a named clinical
decision rule (e.g.\ PERC, Wells, HEART) given a structured input. The
other 23 tools are direct analogues of Corecraft's retail/enterprise
tools.

\subsection{Tool semantics and error model}
\label{sec:semantics}
Tool calls are deterministic given the world state: no latency is
simulated, no retry or timeout semantics are modelled, and no
partial-success or idempotency markers are attached to the mutating
operations. Every call returns either
$\{\texttt{status}{:}\,\texttt{"ok"},\ \texttt{data}{:}\,\ldots\}$
or
$\{\texttt{status}{:}\,\texttt{"error"},\ \texttt{code}{:}\,\ldots,\ \texttt{message}{:}\,\ldots\}$
and is recorded in an append-only audit log that the world state
exposes to the rubric grader. The absence of latency and retry
semantics is deliberate for this release---not because the real world
lacks them, but because modelling them honestly would require a
concurrency story this environment does not yet have
(\S\ref{sec:limits}).

Per Corecraft \S5.5, HealthCraft injects six classes of
structured noise that are seeded deterministically so that evaluations
remain reproducible: (i)~pagination limits (search tools cap at $10$
results with no \texttt{hasMore} flag, exposing the ``failure to
paginate'' pattern from Corecraft \S4.1); (ii)~occasional clock skew
between triage and nursing notes; (iii)~incomplete records (some
patients lack insurance; some encounters lack disposition); (iv)~stale
data (lab results delayed by 2 hours, vitals not yet posted);
(v)~ambiguous documentation (abbreviations, unclear prior notes);
(vi)~red-herring abnormal findings that are irrelevant to the presenting
problem. Noise is part of the environment specification, not of the
task authorship.

\subsection{FHIR integration and OpenEM bridge}
\label{sec:fhir}
The world state stores entities as FHIR R4 resources
(\texttt{src/fhir/resources.py}) generated by the ED-realism FHIR
generator (\texttt{src/fhir/generator.py}), which consumes terminology
bindings (LOINC labs, SNOMED conditions, RxNorm medications) from
\texttt{src/fhir/terminology.py}. Clinical knowledge is indexed through
the OpenEM bridge
(\texttt{src/healthcraft/openem/\{bridge,fhir\_adapter,condition\_mapper\}.py}):
at seed time, OpenEM conditions, confusion pairs, decision rules, and
evaluation properties are loaded and projected onto the entity graph
so that any task grounded in an OpenEM condition has a resolvable
chain all the way to FHIR resources the agent can retrieve.

\subsection{Determinism}
\label{sec:determinism}
Every randomness source is seeded. Entity generation, noise injection,
and task-engine scheduling share a single seed (seed=42)  
set in \texttt{src/healthcraft/world/seed.py}. Model sampling is
temperature=0  
as enforced by the orchestrator
(\texttt{src/healthcraft/llm/orchestrator.py}). Two independent runs
of the same command therefore produce the same trajectories up to
provider-side nondeterminism in the model itself; this is the source
of the $\text{Pass@3} > \text{Pass@1}$ gap we report in
\S\ref{sec:results}.

\section{Reward Function}
\label{sec:reward}

Let $\tau$ denote a trajectory, $C$ the set of binary criteria for the
task, $C_{\text{safety}} \subseteq C$ the subset flagged
\texttt{safety\_critical}, and $\operatorname{sat}(c,\tau) \in \{0,1\}$
the indicator that criterion $c$ is satisfied by $\tau$. Reward is
\begin{equation}
\label{eq:reward}
r(\tau) \;=\;
\begin{cases}
0 & \text{if } \exists\, c \in C_{\text{safety}} : \operatorname{sat}(c, \tau) = 0, \\[2pt]
\dfrac{1}{|C|} \displaystyle\sum_{c \in C} \operatorname{sat}(c, \tau) & \text{otherwise},
\end{cases}
\end{equation}
and a trajectory is counted as \emph{passed} iff every criterion is
satisfied:
\begin{equation}
\label{eq:pass}
\operatorname{Pass}(\tau) \;=\; \mathbf{1}\!\left[\,\forall c \in C:\ \operatorname{sat}(c, \tau) = 1\,\right].
\end{equation}
Equation~\ref{eq:reward} is Corecraft's Eq.~1 augmented with a hard
safety gate: a single unsatisfied safety-critical criterion zeros the
reward regardless of how many other criteria are satisfied. The gate is
the central design choice. Emergency medicine tolerates no tradeoff
between a lethal error and an otherwise competent performance, so the
reward surface must reflect that. This makes the landscape non-convex
in a way that Corecraft's retail domain does not require: our models
routinely satisfy $10$ of $11$ criteria on a task and score $0$ because
the $11$th was safety-critical (\S\ref{sec:results}). Section~4 of the
reference implementation (\texttt{src/healthcraft/tasks/rubrics.py}
lines 55--86) is the source of truth for Eq.~\ref{eq:reward}.

\paragraph{Verification methods.}
Each criterion declares how it is to be verified. The three methods
are:
(i)~\emph{world\_state}, a deterministic check against the append-only
audit log emitted by the MCP layer; a criterion such as ``the agent did
not order anticoagulation'' resolves to a predicate over the audit
entries and is evaluated without an LLM in the loop;
(ii)~\emph{llm\_judge}, a cross-vendor judge (\S\ref{sec:eval}) that
reads the full trajectory and returns a binary decision with evidence;
(iii)~\emph{pattern}, a regex match over the final assistant turn,
retained for light-weight presentational checks (e.g.\ discharge
instructions contain return-precautions language).
We prefer \emph{world\_state} wherever a deterministic predicate is
expressible. Prior work on \ifanon a radiology evaluation harness\else
the RadSlice harness\fi{} motivates this preference: L0 pattern
checks alone had a 75\% false-positive rate; combining them with
state verification eliminated that mode. For HealthCraft, this
choice is particularly consequential because many safety-critical
criteria concern \emph{whether an order was placed}, which is a
well-defined query over the audit log.

\paragraph{Dimensions are diagnostic, not reward.}
The rubric also labels each criterion with one of six weighted
dimensions: \textsf{clinical\_completeness} ($0.20$),
\textsf{clinical\_correctness} ($0.25$),
\textsf{protocol\_adherence} ($0.15$),
\textsf{documentation\_quality} ($0.10$),
\textsf{safety} ($0.20$), and
\textsf{temporal\_sequencing} ($0.10$). These weights do not enter
Eq.~\ref{eq:reward}. They are an analytical lens used only for error
decomposition in post-hoc analysis: when a model fails a task, we want
to know \emph{why}, and grouping criteria by dimension surfaces
patterns (e.g.\ ``the model satisfied completeness but failed
temporal\_sequencing on $80\%$ of multi-step tasks''). Conflating the
weights with the reward would re-introduce the very tradeoff the
safety gate was introduced to prevent.

\section{Task Suite}
\label{sec:tasks}

The public benchmark consists of
195 tasks  
distributed across six categories,  
graded against
2{,}255 binary criteria  
of which
515 are safety-critical.  
Each criterion is a single proposition with a single verification
method; the binary form is load-bearing because it composes cleanly
under Eq.~\ref{eq:reward}. Table~\ref{tab:tasks} summarizes the
distribution; appendix~\ref{app:criteria} decomposes criteria by
dimension.

\begin{table}[t]
\centering
\caption{Task suite. Categories and counts for the V8 release.}
\label{tab:tasks}
\small
\setlength{\tabcolsep}{4pt}
\begin{tabular}{@{}lrrrp{4.4cm}@{}}
\toprule
Category & \# tasks & \# criteria & \# safety-crit. & Generation source \\
\midrule
clinical\_reasoning        & 50 & 639 & 151 & OpenEM confusion pairs (152) \\
multi\_step\_workflows     & 33 & 437 & 109 & Protocol bundles, dispositions \\
clinical\_communication    & 30 & 302 & 35  & Discharge, consult, transfer, MDM \\
safety\_critical\_judgment & 27 & 308 & 137 & OpenEM eval.\ properties (44); \mbox{EMTALA}, capacity \\
information\_retrieval     & 30 & 286 & 37  & OpenEM decision rules (45) \\
temporal\_reasoning        & 25 & 283 & 46  & Overlapping protocols, triage under load \\
\midrule
\textbf{total} & \textbf{195} & \textbf{2{,}255} & \textbf{515} & \\
\bottomrule
\end{tabular}
\end{table}

Tasks carry a difficulty level from $1$ (\emph{Triage}, 1--2 expected
tool calls) to $5$ (\emph{Mass Casualty}, 15+ coordinated calls across
multiple encounters). Difficulty is not weighted into the reward; it
is used for per-level analysis and for sampling during training.

\paragraph{Generation sources.}
Clinical-reasoning tasks are scaffolded from OpenEM's 152 confusion
pairs  
---pairs of conditions whose presentations overlap but whose
management diverges (e.g.\ STEMI vs.\ aortic dissection). The task
asks the agent to disambiguate given FHIR evidence, and the
safety-critical criterion is typically a negated action
(``did not order anticoagulation'' when dissection is in the
differential). Information-retrieval tasks are scaffolded from
OpenEM's 45 decision rules.  
Safety-critical-judgment tasks draw from OpenEM's 44 evaluation
properties  
together with EMTALA and capacity scenarios authored in-house.
Multi-step workflows, clinical communication, and temporal reasoning
are authored categories; each follows a coverage-cycle methodology
adapted from \ifanon a prior emergency-medicine evaluation harness\else
LostBench\fi{} in which gaps are identified by scoring the current
corpus against a target distribution and filling the largest gaps
first.

\paragraph{Release contents.}
Each task ships as a YAML file specifying the description, world
setting (seed, timestamp, active encounter IDs), criteria list, and
metadata. The full task corpus is published alongside the
environment under Apache~2.0.

\section{Evaluation Protocol}
\label{sec:eval}

An evaluation run consists of: (a)~for each of 195 tasks, spin up a
fresh Docker container holding a deterministic world state at
seed=42; (b)~drive the agent for up to $N$ turns (category-dependent)
against the MCP server; (c)~capture the complete audit log and
assistant/tool message trail; (d)~apply the rubric grader to each
trajectory; (e)~aggregate per-model, per-category, and per-task
statistics. Each $(\text{model},\text{task})$ pair is run for
3 trials,  
yielding
585 runs per model  
across the
195 tasks.  
The orchestrator (\texttt{src/healthcraft/llm/orchestrator.py},
function \texttt{run\_frontier\_evaluation}) is idempotent: it
resumes from existing trajectory files on disk, so partial runs can
be continued without re-billing the model.

\paragraph{Metrics.}
Following \textsc{$\tau$-Bench} \citep{tau_bench2024}\ifanon\else{} and
LostBench\fi, we report
\textsc{Pass@1} (mean pass rate across all trials),
\textsc{Pass@3} (at least one of three trials passes, best-case
reliability),
\textsc{Pass$^3$} (all three trials pass, worst-case reliability), the
expected reward $\mathbb{E}[r(\tau)]$ under Eq.~\ref{eq:reward}, and
the safety-failure rate (fraction of trials with at least one
unsatisfied safety-critical criterion). Wilson score 95\% confidence
intervals are reported for all proportions; the derivation is in
Appendix~\ref{app:wilson}.

\paragraph{Cross-vendor judging.}
Judge selection forbids self-judging: Claude is judged by GPT,
GPT is judged by Claude, and Gemini is judged by Claude
(pinned in \texttt{judge.py}). Temperature is held at $0$
for both agent and judge; the judge sees the full trajectory and
returns a JSON-structured verdict per criterion. Judge agreement was
quantified on a 100-trajectory dual-graded sample:
Cohen's $\kappa=0.553$,  
moderate agreement by Landis--Koch. As a partial mitigation we
maintain an oracle validation on task CC-001, where all five oracle
verdicts agree with the \texttt{world\_state} verification method
(5/5);  
this confirms the judge is not drifting on the
\texttt{world\_state}-verifiable subset. The $\kappa=0.553$ figure
remains a real limitation (\S\ref{sec:limits}); we do not claim
human-level judge reliability. As a second, independent test, we built
a deterministic world\_state overlay for 44 llm\_judge criteria
(Appendix~\ref{app:v9-overlay}), re-graded every cached V8 trajectory
without an LLM, and labelled each of the 63 resulting disagreements
under a six-class taxonomy. Agreement is 76.1\%,  
overall $\kappa$ is 0.402,  
73\% of disagreements  
are judge hallucinations (false-positive PASSes that cite tool calls
absent from the audit log), and only $\sim 8\%$ are overlay faults;
this fixes a pointwise upper bound on judge reliability at the
\texttt{llm\_judge}-graded subset.

\paragraph{Models.}
Main results in \S\ref{sec:results} are reported for Claude Opus 4.6
and GPT-5.4. Gemini 3.1 Pro is in flight for the V9 release and
will be appended in a follow-up; the infrastructure supports it
unchanged (cross-vendor judge rules extend naturally).

\section{Results}
\label{sec:results}

Table~\ref{tab:main} reports V8 results for Claude Opus~4.6 and
GPT-5.4. Under Eq.~\ref{eq:reward}, Claude achieves Pass@1 of
24.8\% (Wilson 95\% CI $[21.5, 28.4]$)  
and GPT-5.4 achieves 12.6\% $[10.2, 15.6]$.  
Pass@3 is
37.9\% for Claude  
and
24.6\% for GPT-5.4;  
Pass$^3$ is
13.8\% for Claude  
and
3.1\% for GPT-5.4,  
meaning that even the stronger model passes every one of three
independent trials on only about one task in seven, and that
Pass$^k$ does not yet meet the $\ge 95\%$ deployment gate used
elsewhere in this research programme. Expected reward is
$0.634$ for Claude  
and
$0.546$ for GPT-5.4.  
Safety-failure rates are
27.5\% $[24.1, 31.3]$ for Claude  
and
34.0\% $[30.3, 37.9]$ for GPT-5.4  
---in the latter case, roughly one trial in three contains at least
one unsatisfied safety-critical criterion.

\begin{table}[t]
\centering
\caption{V8 main results. 195 tasks, 3 trials per (model,task). Wilson
95\% confidence intervals in brackets. Gemini~3.1~Pro pending V9
completion.}
\label{tab:main}
\small
\begin{tabular}{lrrrrr}
\toprule
Model & Pass@1 & Pass@3 & Pass$^3$ & Avg.\ reward & Safety fail \\
\midrule
Claude Opus 4.6 & 24.8\% $[21.5, 28.4]$ & 37.9\% & 13.8\% & $0.634$ & 27.5\% $[24.1, 31.3]$ \\
GPT-5.4         & 12.6\% $[10.2, 15.6]$ & 24.6\% & 3.1\%  & $0.546$ & 34.0\% $[30.3, 37.9]$ \\
\bottomrule
\end{tabular}
\end{table}

\begin{figure}[t]
\centering
\includegraphics[width=0.9\linewidth]{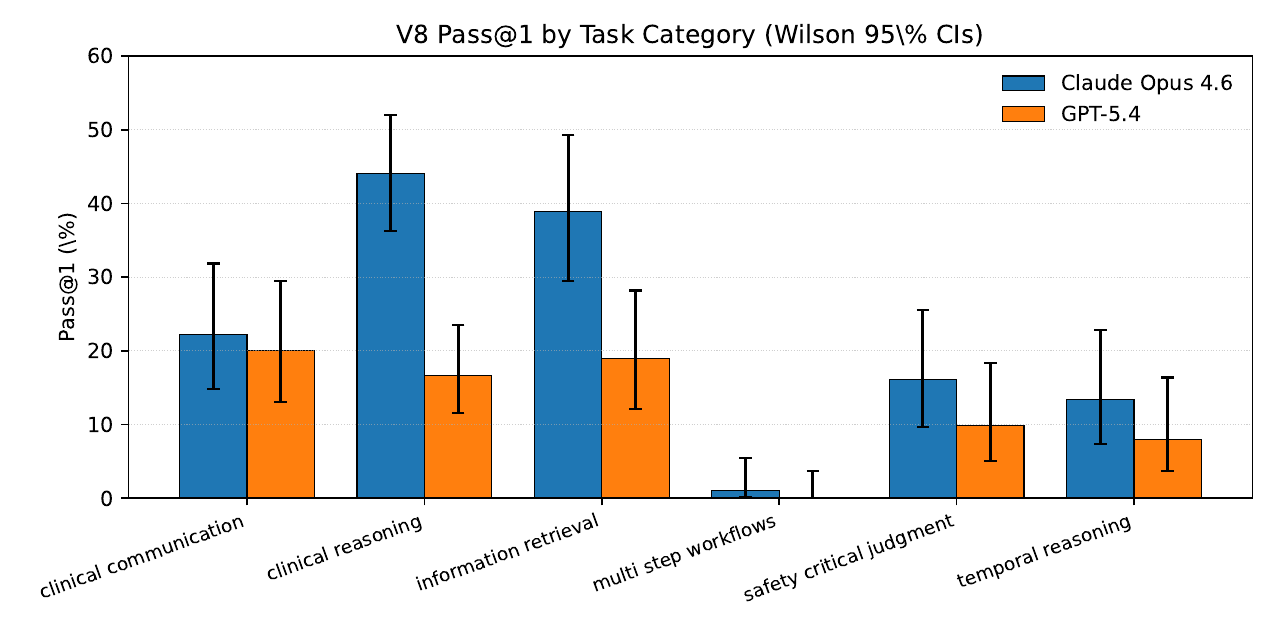}
\caption{V8 Pass@1 by task category with Wilson 95\% CIs.}
\label{fig:percategory}
\end{figure}

\begin{figure}[t]
\centering
\includegraphics[width=\linewidth]{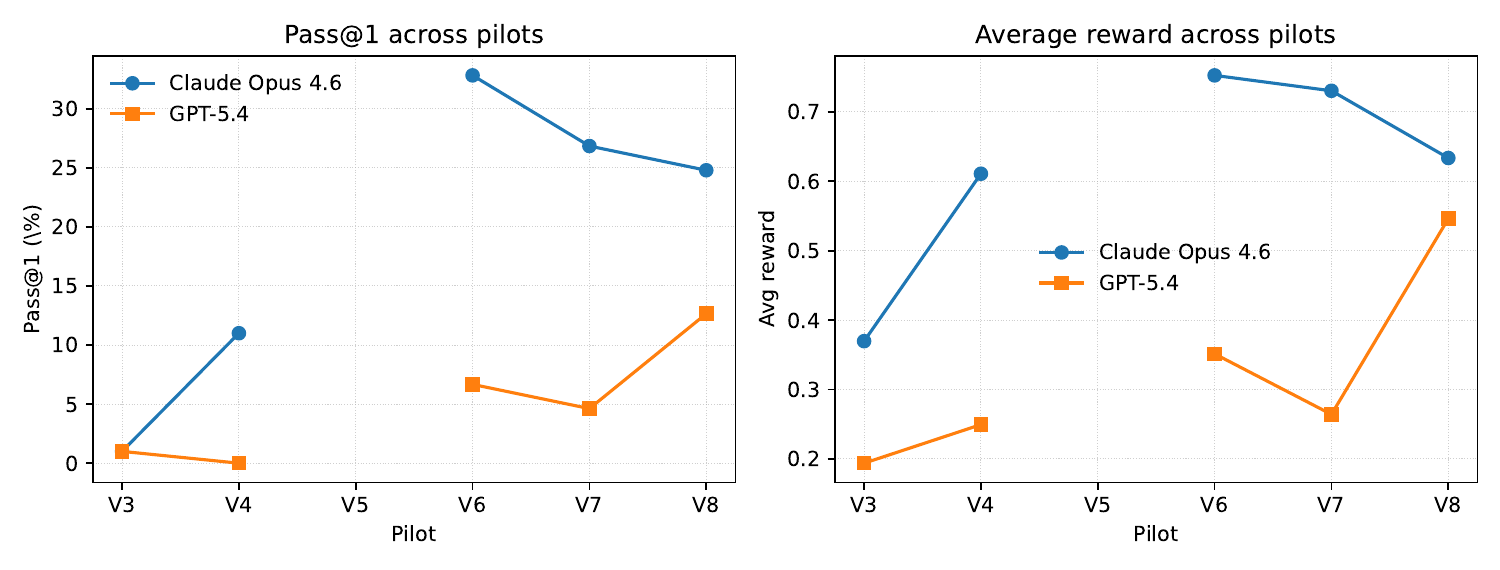}
\caption{Pass@1 and mean reward across pilots v3--v8.}
\label{fig:progression}
\end{figure}

Figure~\ref{fig:percategory} shows per-category Pass@1 with Wilson
95\% error bars. Figure~\ref{fig:progression} plots Pass@1 and mean
reward across pilots v3--v8. Figure~\ref{fig:gate} is a per-task
scatter of mean reward against safety-gate pass fraction and makes
the safety-gate-dominance pattern visible.

\subsection{The multi\_step\_workflows collapse}
\label{sec:msw-collapse}
\begin{figure}[t]
\centering
\includegraphics[width=0.85\linewidth]{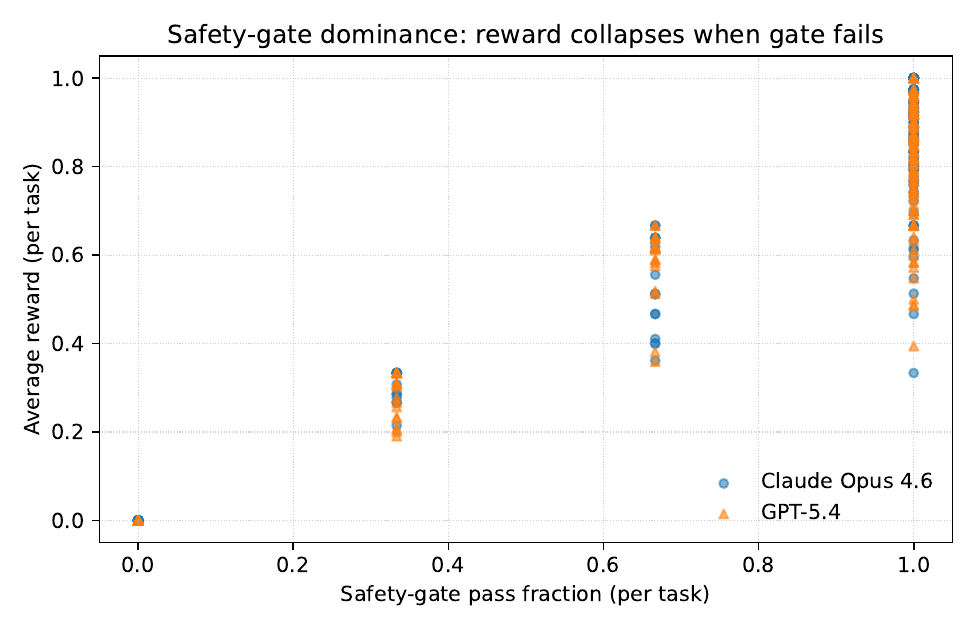}
\caption{Per-task safety-gate dominance. Each point is one task. The
band along the top-left (high safety-gate pass, high reward) is the
expected regime; the vertical band along reward $\approx 0$ for tasks
with safety-gate fraction below $\sim 0.5$ is the
\texttt{multi\_step\_workflows} collapse.}
\label{fig:gate}
\end{figure}

Both models collapse on \texttt{multi\_step\_workflows}: Claude passes
1.0\% of tasks  
and GPT-5.4 passes
0\% of tasks.  
Twenty-one of the 33 tasks in this category see zero passes across 3
trials by either model. The mechanism is visible in
Fig.~\ref{fig:gate}: on these tasks the agents frequently satisfy
most non-safety criteria (mean raw criterion satisfaction is in
the $0.4$--$0.6$ band) but have near-zero safety-gate pass rates.
The reward surface collapses to $0$ not because the agents are unable
to perform the workflow, but because the workflow touches enough
safety-critical decision points---transfer checks, prior-auth
verification, explicit EMTALA acknowledgement---that one slip zeros
the entire trial.

\subsection{Methodological transparency: superseded V7}
\label{sec:v7-superseded}

Between pilots v2 and v8 we fixed
six infrastructure bugs  
in the harness---none of them in the agents, the judge, or the
rubric, and none of them known at the time of the v7 results that
we had previously circulated. The six bugs were:
(i)~the evaluator silently dropped non-terminal qualifiers when
parsing tool arguments, preventing some safety-critical criteria from
ever being satisfiable;
(ii)~tool-schema mismatches between the server-side JSON Schema and
the agent-visible MCP schema caused confident tool calls to fail
validation on previously valid input;
(iii)~entity-injection gaps left some FHIR bundles missing
\textsc{ServiceRequest} links that certain tasks' criteria required;
(iv)~system-prompt composition double-appended policy blocks in
certain task variants, pushing relevant clinical context out of the
token budget;
(v)~OpenEM loading had a fallback path that silently degraded to five
bundled conditions when the full corpus was unavailable; and
(vi)~\texttt{processTransfer} had an ordering bug that marked a
transfer complete before the receiving-facility handshake was
recorded in the audit log. V8 resolves all six; the corresponding
PR chain is logged in \texttt{docs/V8\_ANALYSIS.md} and
\texttt{docs/EVALUATION\_INTEGRITY.md}.

Offline re-scoring of V7 trajectories under the V8 evaluator produced
611 criterion-level verdict flips overall (602 satisfied-to-unsatisfied
from the qualifier-enforcement fix; 9 unsatisfied-to-satisfied from
the compound-clause fix) with zero additional flips under the V8
evaluator applied to V8 trajectories, which we take as a lower bound
on evaluator stability. The three tool-side bugs
(\texttt{processTransfer} $\times 2$ plus the missing
entity-ID injection) primarily deflated GPT-5.4 because GPT follows
schema parameter names literally while Claude adapts to error
messages; this asymmetry is the mechanism behind the qualitative
re-ordering reported below. We did not run a controlled ablation of
each bug in isolation because the six fixes were landed as one PR
chain against a single rollout harness; re-instrumenting v7 to
toggle bugs individually was judged more likely to introduce new
artefacts than to cleanly decompose the observed delta.

The pilot deltas are instructive. On average reward, Claude Opus~4.6
\emph{drops} from v7 to v8 by about $13\%$ relative  
(0.730$\to$0.634) while GPT-5.4 \emph{rises} by about $107\%$ relative  
(0.264$\to$0.546). Pass@1 follows the same directions with smaller
magnitudes (Claude 26.8\%$\to$24.8\%, GPT 4.6\%$\to$12.6\%). We are
cautious about assigning causation---the six bug fixes were committed
in a single chain and their isolated effects were not ablated---but the
direction of the flip is consistent across metrics. We highlight the
transparency here because the alternative (treating the v7 numbers as
authoritative, or burying the correction in a footnote) would have
meant publishing a result whose headline ordering is at least partly
an artefact of harness bugs. The broader lesson is that for tool-use
benchmarks, infrastructure fidelity is not a separate axis from model
evaluation; it is part of what is being measured.

\section{Limitations and Future Work}
\label{sec:limits}

We describe the real limits of this release directly. None of them
invalidates the results reported above; all of them constrain what
those results can be used to claim.

\paragraph{Scope and non-claims.}
HealthCraft measures \emph{execution correctness} under a fixed
rubric in a deterministic FHIR tool-use environment; it does not
measure clinical reasoning under uncertainty, calibrated risk
judgement, or deployment-time behaviour. Safety is operationalized as
binary criterion satisfaction with a hard gate, not as dynamic risk
tradeoff management. Determinism---fixed seed, temperature zero,
static patient state---is a methodological requirement for
reproducible evaluation, distinct from the stochastic variability
that deployment confronts. FHIR~R4 grounding is a claim about
interoperability surfaces, not about team communication,
interruptions, or the social dynamics of an emergency department. A
strong result on HealthCraft is necessary but not sufficient evidence
of clinical readiness, and we state this so downstream use of the
benchmark does not overclaim.

\paragraph{Static patient state.}
Between agent actions, patient state does not evolve. Vitals do not
trend, the septic patient does not decompensate, and the triaged-low
patient does not bounce back to triage. Emergency medicine is a
temporal discipline, and this is the largest single fidelity gap in
the environment. A future release will introduce a patient state
machine whose transitions are driven by a combination of elapsed
simulated time and the agent's actions.

\paragraph{No latency, retry, or timeout semantics.}
Tool calls return instantaneously and deterministically. Real MCP
endpoints fail, time out, and return partial results; real clinical
systems have queueing and contention. We chose not to simulate these
rather than simulate them badly; the present environment should be
treated as an upper bound on the agent's performance under ideal
infrastructure. Introducing honest latency requires a concurrency
model (e.g.\ simulated wall-clock with action budgets) that we are
not shipping in this release.

\paragraph{No idempotency markers on mutating operations.}
Wave~3 tools do not accept an idempotency key, so a retried
\texttt{createClinicalOrder} can silently create a duplicate order.
Our tasks are structured so this is not exercised, but an RL loop
that encourages retries would need this before being safe to train
against.

\paragraph{LLM-judge reliability.}
Cohen's $\kappa = 0.553$ is moderate, not substantial, and our
oracle-validation bridge (\S\ref{sec:eval}) covers only the
\texttt{world\_state}-verifiable subset of criteria. For criteria
graded by \texttt{llm\_judge}, there is real residual variance that
our intervals do not capture. Stratified by trajectory length
(appendix~\ref{app:judge}), $\kappa$ is lowest in the long-trajectory
regime ($>40$ turns, $\kappa=0.306$) where raw agreement is in fact
highest ($92.0\%$); this base-rate paradox matters for Claude
(median $\sim 49$ turns) more than for GPT-5.4 (median $\sim 8$). The
deterministic overlay audit (Appendix~\ref{app:v9-overlay}) sharpens
this: of 63 disagreements between the overlay and the V8 judge,
46 are judge hallucinations  
(73\% of disagreements), and 5 of the 6 safety-critical
inversions  
are judge PASS verdicts on trajectories in which the audit log
contains no successful qualifying tool call---i.e.\ the judge
invented the passing behaviour. Under the Tier~1 subset of 10 overlay
entries  
where disagreements are exhausted by judge hallucination plus
$\le 2$ safe labels, we recommend replacing \texttt{llm\_judge} with
the deterministic rule for reward; Tier~2 (26 entries,  
research-only) and Tier~3 (8 entries,  
keep \texttt{llm\_judge}) retain the LLM grader. The partial mitigation
is to prefer \texttt{world\_state} at authoring time and to apply
Tier~1 overlays where available; the full mitigation is
physician-in-the-loop adjudication, which is underway on a related
corpus and will be ported to HealthCraft criteria in a follow-up.

\paragraph{Infrastructure-bug isolation.}
The V7-to-V8 delta (\S\ref{sec:v7-superseded}) is reported as a bundle
of six simultaneous fixes. We did not ablate individual bugs; the
qualitative re-ordering of Claude and GPT-5.4 is therefore a
\emph{joint} effect of the six, not attributable to any single one.
Per-bug direction of impact is recorded in
\texttt{docs/EVALUATION\_INTEGRITY.md}. Independent replication of v7
under the v8 evaluator would be the natural next step for
external validation of the fixed numbers.

\paragraph{Task coverage.}
The 195 tasks sample from OpenEM's 370 conditions at an average of
under one task per condition. There are known coverage gaps
(paediatric subspecialties, some toxicology presentations, most
obstetric emergencies); the coverage-cycle methodology makes them
visible but does not close them in this release.

\paragraph{Task-distribution balance and the negative-class slate.}
The V8 slate over-samples action-positive scenarios where
the correct behaviour is to order, escalate, or admit and
under-samples action-negative ones where the correct
behaviour is restraint. As a shipped structural fix we
committed a $10$-task 
negative-class slate (NEG-001 through NEG-010,
Appendix~\ref{app:v9-overlay}) adding $82$ criteria--
$14$ safety-critical-
for a total of $205$ tasks 
and $2{,}337$ criteria. 
A 60-run smoke pilot (\texttt{docs/NEG\_SMOKE\_PILOT.md})
establishes that the slate rebalances \emph{task-level}
difficulty: mean Pass@1 is $0.500$ 
for Claude Opus 4.7 and $0.533$ 
for GPT-5.4, versus V8 corpus means of $0.248$ and $0.126$.
It also caught two iatrogenic patterns the V8 corpus does
not surface: insulin infusion on alcoholic ketoacidosis
($4/6$ trials across both models) 
and tPA on hypoglycaemic aphasia
($2/6$ trials). 
The pilot also \emph{falsifies} the Appendix F.5 projection
that NEG criteria would average ${\sim}0.65$ criterion-PASS
and pull audit-subset prevalence into the target band:
observed NEG-criterion prevalence is $0.929$. 
Restraint criteria are trivially satisfied whenever the
agent has been trained against the specific over-action the
criterion forbids, so the kappa-paradox mitigation that
remains open is physician adjudication of the Tier~2 subset
(Appendix~\ref{app:v9-overlay}), not further task authoring.

\paragraph{Evaluation-to-training boundary.}
The same Eq.~1 signal that supports evaluation is the nominal reward
an RL loop would read. The NEG smoke-pilot result above is direct
empirical evidence that it is not drop-in training-safe: a
training loop that credited restraint criteria at their observed
PASS prevalence of $0.929$ 
would be pushed toward pattern-matching
the specific over-actions frontier models have already been trained
against, rather than toward generalized clinical judgement. We
scaffold the coupling to Megatron+SGLang+GRPO per
Corecraft~\S5.2 but do not claim training-safety; closing this
boundary requires a soft-gate/hard-gate ablation, restraint-criterion
reweighting, and reward-hacking probes that we leave as future work.

\paragraph{Underpowered for scaling-law claims.}
Two-model main results (Gemini~3.1~Pro pending) are diagnostic but
underpowered for scaling-law claims. Corecraft Table~1 reports six
frontier configurations; we cite it where broader context is needed
and defer our own scaling analysis.

\paragraph{Future work.}
Dynamic patient state; full RL training integration (Megatron +
SGLang + GRPO per Corecraft \S5.2); task-set scaling to a
$\sim\!1000$-task training split kept disjoint from the
evaluation split; physician-adjudication data on HealthCraft
criteria; the V10 pilot that will re-grade against the
post-NEG $205$-task corpus.

\section{Conclusion}
\label{sec:conclusion}

HealthCraft is a reinforcement-learning environment for emergency
medicine built to be honest about what it measures. The FHIR-R4
world state and 24-tool MCP surface are grounded in an interoperability
standard used in production clinical systems; the dual-layer rubric
with its hard safety gate reflects the way emergency medicine
actually weighs tradeoffs; and the 195-task benchmark is traceable
back to OpenEM's peer-reviewed clinical sources.

The V8 results are not triumphant. Frontier models pass
one task in four (Claude) or one in eight (GPT-5.4) at
Pass@1. One trial in three contains a safety-critical violation.
Multi-step workflows collapse almost completely. These are not the
numbers a deployment decision would be based on, and that is the
point: the environment's job is to make those numbers visible, with
confidence intervals, so that the deployment conversation can be
honest.

The harness, the world-state seed, the 24-tool server, the 195 tasks
and their 2{,}255 criteria, the rubric, the cross-vendor judge rules,
and the Docker bundle are all released under Apache~2.0
\ifanon at a URL redacted for blind review\else at
\url{https://github.com/GOATnote-Inc/healthcraft}\fi. The environment
scaffolds coupling to a Megatron+SGLang+GRPO training loop per
Corecraft~\S5.2; training-reward ablations are future work given
the restraint-prevalence finding in Appendix~\ref{app:v9-overlay}
and the Evaluation-to-training boundary in \S\ref{sec:limits}. We encourage
others to break the environment, extend the task suite, and report
results.


\bibliographystyle{plainnat}
\bibliography{references}

\appendix

\section{Full MCP tool schema}
\label{app:tools}

Table~\ref{tab:tools} gives the complete list of the 24~MCP tools. The
canonical JSON Schema (with per-argument constraints, return types, and
error codes) lives in \texttt{configs/mcp-tools.json} and is loaded
directly by the FastMCP server.

\begin{table}[h]
\centering
\caption{MCP tool surface. Tool names are camelCase at the MCP
boundary and snake\_case internally.}
\label{tab:tools}
\small
\begin{tabular}{lll}
\toprule
Wave & Type & Tools \\
\midrule
1 & read
    & \texttt{searchEncounters}, \texttt{searchPatients},
      \texttt{searchClinicalKnowledge}, \\
 & & \texttt{searchReferenceMaterials}, \texttt{searchAvailableResources}, \\
 & & \texttt{getEncounterDetails}, \texttt{getConditionDetails},
      \texttt{getPatientHistory}, \\
 & & \texttt{getProtocolDetails}, \texttt{getTransferStatus},
      \texttt{getInsuranceCoverage}, \\
 & & \texttt{getReferenceArticle} \\
\addlinespace
2 & compute
    & \texttt{checkResourceAvailability},
      \texttt{calculateTransferTime}, \\
 & & \texttt{runDecisionRule}, \texttt{validateTreatmentPlan} \\
\addlinespace
3 & mutate
    & \texttt{createClinicalOrder}, \texttt{updateTaskStatus},
      \texttt{updateEncounter}, \\
 & & \texttt{updatePatientRecord}, \texttt{registerPatient},
      \texttt{applyProtocol} \\
\addlinespace
4 & workflow & \texttt{processDischarge}, \texttt{processTransfer} \\
\bottomrule
\end{tabular}
\end{table}

All wave-1 search tools cap results at 10 with no \texttt{hasMore}
signal. All wave-3 and wave-4 tools validate inputs before applying
the mutation and log structured audit entries. Error responses are
uniform:
$\{\texttt{status}{:}\,\texttt{"error"},\ \texttt{code}{:}\,\ldots,\ \texttt{message}{:}\,\ldots\}$.

\section{Criteria by dimension}
\label{app:criteria}

The 2{,}255 binary criteria decompose approximately as follows
across the six weighted dimensions, aggregated over all task
categories: \textsf{clinical\_completeness}~$\sim$20\%;
\textsf{clinical\_correctness}~$\sim$25\%;
\textsf{protocol\_adherence}~$\sim$15\%;
\textsf{documentation\_quality}~$\sim$10\%;
\textsf{safety}~$\sim$20\% (the 515 safety-critical criteria are a
superset of the \textsf{safety} dimension and can additionally appear
in any of the other dimensions when they were authored with
safety\_critical=true);
\textsf{temporal\_sequencing}~$\sim$10\%. Exact counts per category
are emitted by \texttt{scripts/analyze\_results.py}. Dimension weights
are used only for post-hoc error decomposition; they do not enter
Eq.~\ref{eq:reward}.

\section{Pilot v2--v8 changelog}
\label{app:pilots}

The six infrastructure bugs described in \S\ref{sec:v7-superseded} were
fixed across pilots v6, v7, and v8. Full per-task pass/fail tables for
every pilot live in \texttt{results/pilot-v\{3..8\}-\{claude-opus,gpt54\}/}.
Pilot-level deltas for Pass@1 and mean reward across the v3--v8 run
are plotted in Figure~\ref{fig:progression}.

\section{Wilson score confidence interval derivation}
\label{app:wilson}

For $k$ successes out of $n$ trials, the Wilson score 95\% CI is
\[
\left(\frac{\hat p + \frac{z^2}{2n}}{1 + z^2/n}\right)
\pm
\frac{z\sqrt{\frac{\hat p(1-\hat p)}{n} + \frac{z^2}{4n^2}}}{1 + z^2/n},
\]
with $\hat p = k/n$ and $z = 1.96$. Wilson is preferred over the
normal approximation when $n \hat p (1 - \hat p)$ is small; over
Clopper--Pearson it is less conservative without becoming
anti-conservative in the $n\sim 200$ regime relevant for this
benchmark. The implementation is
\texttt{src/metrics/confidence\_intervals.py}.

\section{Judge reliability details}
\label{app:judge}

Cross-vendor assignment is pinned by
\texttt{src/healthcraft/llm/judge.py}:
Claude Opus~4.6 $\to$ GPT-5.4 judge; GPT-5.4 $\to$ Claude Opus~4.6
judge; Gemini~3.1~Pro $\to$ Claude Opus~4.6 judge. Cohen's
$\kappa=0.553$ was computed over a 100-criterion test-retest sample
(each criterion graded three times, $n=300$ calls), with 77.0\% raw
self-agreement. Oracle validation on task CC-001 returned 5/5 agreement
with the \texttt{world\_state} verification method, indicating that for
the \texttt{world\_state}-verifiable subset the judge does not
introduce additional drift.

\paragraph{Stratification by trajectory length.}
Agreement and $\kappa$ were stratified by trajectory length: short
trajectories ($<15$ turns, $n=5$) showed 60.0\% agreement /
$\kappa=0.444$; medium ($15$--$40$ turns, $n=70$) 72.9\% / $0.542$;
long ($>40$ turns, $n=25$) 92.0\% / $0.306$. The long-trajectory
stratum is the paradoxical case: raw agreement is highest there but
$\kappa$ is lowest. This is Cohen's kappa's well-known base-rate
artefact: when agreement is near the ceiling (the judge almost
always returns ``satisfied'' on long trajectories because those
trajectories tend to contain the evidence the criterion is looking
for), the chance-agreement denominator grows faster than the observed
agreement numerator and $\kappa$ collapses. We report both statistics
because they answer different questions: raw agreement is the rate
at which re-grading would flip a verdict, while $\kappa$ is the rate
above chance. The long-trajectory regime is more relevant for Claude
(median $\sim 49$ turns) than GPT-5.4 (median $\sim 8$ turns), which
is why the judge-reliability concern in \S\ref{sec:limits} is framed
as a bound on Claude-favouring variance rather than symmetric noise.

\section{V9 deterministic overlay audit of the LLM judge}
\label{app:v9-overlay}

Cross-vendor judging (\S\ref{sec:eval}) does not eliminate judge
variance; Cohen's $\kappa=0.553$ leaves appreciable residual
disagreement. Because the judge is the unit of measurement for
63.5\% of criteria 
(\S\ref{sec:limits}), its noise floor bounds every downstream claim.
We therefore built an independent deterministic audit:
44 criteria 
originally graded by \texttt{llm\_judge} were migrated to
\texttt{world\_state} predicates and re-graded against the cached V8
trajectories. This appendix reports what that audit found; the full
per-observation report is committed at
\texttt{docs/V9\_OVERLAY\_AUDIT.json}.

\paragraph{Overlay construction.}
The migration pipeline has three stages. (i)~Auto-migration:
\texttt{scripts/migrate\_criteria.py} scans every
\texttt{llm\_judge} assertion for patterns that cleanly map onto
audit-log predicates (e.g.\ ``agent ordered $X$'' $\mapsto$
``audit\_log contains \texttt{createClinicalOrder} with
medication$=X$''). (ii)~Confidence filtering:
\texttt{scripts/populate\_v9\_overlay.py} retains only
\texttt{confidence == 'high'} candidates and flags
subordinate-clause assertions for review.
(iii)~Manual correction of a small number of wrong-entity
mappings (e.g.\ an overlay row whose assertion concerned blood
cultures was originally wired to a blood-product tool). The
resulting overlay
(\texttt{configs/rubrics/v9\_deterministic\_overlay.yaml}) has
$44$ entries across five categories.

\paragraph{The \texttt{contains attempt at} directive and the
\texttt{intent\_rescue\_reason} contract.}
Many safety-critical criteria are phrased in terms of
\emph{what the agent attempted} rather than what succeeded.
A simulator bug in V8 prevented the \texttt{blood\_admin}
task type from executing, so well-formed orders for blood
products returned
$\{\texttt{status:error},\ \texttt{code:unknown\_task\_type}\}$
in the audit log. The V8 LLM judge correctly scored these as
agent intent (PASS); a naive world-state check would score
them as FAIL. To preserve this semantics deterministically
we added a \texttt{contains attempt at} directive to the
evaluator. An early implementation accepted \emph{any} audit
entry matching the tool/params filter regardless of
\texttt{result\_summary}, which turned out to be
over-permissive (\S F.4 below). The production directive
accepts an audit entry as evidence of attempt only if
\texttt{result\_summary == "ok"}, or
\texttt{result\_summary == "error"} with an
\texttt{error\_code} in a fixed simulator-side allowlist:
$\{\texttt{unknown\_task\_type},$
$\texttt{not\_implemented},$
$\texttt{simulator\_error},$
$\texttt{internal\_error},$
$\texttt{service\_unavailable}\}$.
Agent-side schema errors (\texttt{missing\_param},
\texttt{invalid\_params}, \texttt{unknown\_tool}) are not
valid attempts. In addition, every overlay entry using
\texttt{contains attempt at} must declare a free-text
\texttt{intent\_rescue\_reason} in the overlay YAML; the
orchestrator refuses to load the overlay if any entry is
missing this attestation. The attestation is the design
pattern; the allowlist is its enforcement.

\paragraph{Forensic walkthrough: SCJ-005-C04.}
A pre-tightening run of the overlay classified two trials
of SCJ-005-C04 (GPT-5.4 seeds $42$ and $44$) as
$\text{v9}=\text{PASS}$ where V8 had marked them FAIL, with
the disagreement classifier assigning the label
\texttt{infrastructure\_error}. On direct audit, however,
only one trial ($\_44\_t3$) carried a genuine sim-side
\texttt{unknown\_task\_type}. The other ($\_42\_t1$) contained
$12/13$ tool errors of type \texttt{missing\_param}: the
agent had omitted the required \texttt{details} field from
its \texttt{createClinicalOrder} calls and never once
produced a well-formed request. Under a principled reading
of ``attempted'' this trial had not attempted the order, it
had emitted malformed calls. Under the tightened directive,
$\_42\_t1$ correctly flips back to $\text{v9}=\text{FAIL}$
while $\_44\_t3$ retains its rescue. The pre-tightening
overlay was laundering an agent-side schema failure as a
clinical attempt. We report the walkthrough because the
failure mode---a naive intent check that accepts any error
code as evidence of intent---is easy to recreate, and the
fix (the sim-side allowlist plus per-entry attestation) is
narrow enough to be reused.

\paragraph{Agreement.}
Agreement was computed on
$n=264$ 
(criterion $\times$ trial) observations: $44$ overlay
criteria $\times$ $2$ models (Claude Opus 4.6, GPT-5.4)
$\times$ $3$ trials. The V8 judge verdict is the reference;
the V9 overlay verdict is the test. Overall raw agreement is
$76.1\%$ 
with Cohen's
$\kappa = 0.402$ 
and
$6$ 
safety-critical verdict inversions. Per-category agreement
is given in Table~\ref{tab:v9_cat_agreement}.

\begin{table}[h]
\centering
\caption{V9 overlay vs.\ V8 judge agreement by category.
$n$ is (criteria~$\times$~trials). $\kappa$ is Cohen's;
PABAK is prevalence-and-bias-adjusted kappa. V8 PASS
prevalence is high across all categories, which is why
$\kappa$ is systematically lower than PABAK (the
$\kappa\ge 0.80$ discussion below).}
\label{tab:v9_cat_agreement}
\small
\begin{tabular}{lrrrrr}
\toprule
Category & $n$ & Agreement & V8 prev. & $\kappa$ & PABAK \\
\midrule
clinical\_communication     & $12$  & $58.3\%$ & $58.3\%$ & $0.211$ & $0.167$ \\
clinical\_reasoning         & $12$  & $75.0\%$ & $66.7\%$ & $0.526$ & $0.500$ \\
multi\_step\_workflows      & $78$  & $82.1\%$ & $78.2\%$ & $0.533$ & $0.641$ \\
safety\_critical\_judgment  & $54$  & $77.8\%$ & $83.3\%$ & $0.265$ & $0.556$ \\
temporal\_reasoning         & $108$ & $73.1\%$ & $82.4\%$ & $0.335$ & $0.463$ \\
\midrule
all                         & $264$ & $76.1\%$ & $79.5\%$ & $0.402$ & $0.523$ \\
\bottomrule
\end{tabular}
\end{table}

\paragraph{Disagreement taxonomy.}
A rule-based classifier
(\texttt{scripts/kappa\_validation.py::\_classify\_disagreement})
assigns one of six labels to each disagreeing observation in
priority order: \texttt{overlay\_wrong\_entity} (overlay
targets wrong tool/params); \texttt{infrastructure\_error}
(v9 PASS rescued by a sim-side error);
\texttt{judge\_hallucination} (v8 PASS with zero successful
matching tool calls in trajectory);
\texttt{intent\_execution\_split} (agent attempted but did
not complete; judge scored intent);
\texttt{vocab\_gap} (medication-class terminology mismatch);
and \texttt{conditional\_logic} (assertion has a conditional
clause the overlay cannot encode). Counts for $63$
disagreeing observations are in Table~\ref{tab:v9_labels}.
The dominant class is \texttt{judge\_hallucination} at
$73\%$ 
($46$ of $63$): 
V8 marked the criterion PASS yet the trajectory contains
zero successful matching tool calls in the audit log. These
are judge false positives, and because most HealthCraft
safety-critical criteria concern whether a tool was invoked,
they are a direct ceiling on judge reliability.

\begin{table}[h]
\centering
\caption{Disagreement classifier counts over the $63$
disagreeing observations. Safety column counts the
safety-critical subset.}
\label{tab:v9_labels}
\small
\begin{tabular}{lrrr}
\toprule
Label & Count & \% of disagreements & Safety-critical \\
\midrule
judge\_hallucination       & $46$ & $73\%$ & $5$ \\
infrastructure\_error      & $5$  
                                   & $8\%$  & $1$ \\
vocab\_gap                 & $5$  & $8\%$  & $0$ \\
intent\_execution\_split   & $4$  & $6\%$  & $0$ \\
overlay\_wrong\_entity     & $2$  & $3\%$  & $0$ \\
conditional\_logic         & $1$  & $2\%$  & $0$ \\
\midrule
total                      & $63$ & $100\%$ & $6$ \\
\bottomrule
\end{tabular}
\end{table}

\paragraph{Tiering.}
On the basis of the disagreement taxonomy each of the 44
overlay entries is assigned one of three tiers:

\begin{itemize}
  \item \textbf{Tier 1 (reward-safe): $10$ entries.} 
  All six trials agree with the V8 judge. The overlay can
  replace \texttt{llm\_judge} for these criteria in reward
  computation without changing any verdict.
  \item \textbf{Tier 2 (research-only): $26$ entries.} 
  Disagreement labels are a subset of $\{$\texttt{judge\_hallucination},
  \texttt{infrastructure\_error}$\}$. The overlay is
  plausibly correct and the judge is plausibly at fault,
  but we do not use the overlay for reward until physician
  adjudication has confirmed the direction.
  \item \textbf{Tier 3 (keep \texttt{llm\_judge}):
  $8$ entries.} 
  At least one disagreement carries a label outside the
  judge-noise set (\texttt{overlay\_wrong\_entity},
  \texttt{intent\_execution\_split}, \texttt{vocab\_gap},
  \texttt{conditional\_logic}). These flag real overlay or
  semantic issues; the V8 judge remains binding until the
  overlay is corrected.
\end{itemize}

The tiering is presented not as a HealthCraft-specific
artefact but as a reusable template: given any
deterministic-overlay audit of an LLM judge, the six-label
classifier over disagreements induces a three-way tiering
(reward-safe~/ research-only~/ keep-judge), and only the
reward-safe tier changes the production reward signal.

\paragraph{Why $\kappa \ge 0.80$ is the wrong gate here.}
Cohen's $\kappa$ treats the reference grader as ground
truth. The disagreement taxonomy shows that treatment is
unsafe for this audit: $73\%$ of observed disagreements
trace to judge false positives, not overlay faults. A
perfect overlay audited against a judge with
$\sim 23\%$ false-positive rate on the audited subset
cannot reach $\kappa \ge 0.80$. This is an instance of the
kappa paradox in the high-prevalence regime (V8 PASS
prevalence
$79.5\%$ 
on the audited subset): agreement is dominated by the
shared positive class, and the chance-agreement
denominator grows faster than the observed-agreement
numerator. The honest statistic for
\emph{reliability-of-overlay-given-noisy-judge} is raw
agreement paired with the disagreement taxonomy; the
honest statistic for
\emph{reliability-of-judge-given-physician-adjudicated-truth}
is $\kappa$ against an adjudicated subset, which is
underway in a follow-up.

\paragraph{Prevalence correction: negative-class task slate.}
\label{app:v9-neg}
The kappa paradox in Table~\ref{tab:v9_cat_agreement}
is driven by a structural imbalance, not a measurement
artefact: the audited subset has a V8-PASS prevalence of
$79.5\%$, 
so even a well-calibrated overlay cannot reach
$\kappa\ge 0.80$ without first re-balancing what the
benchmark measures. The V8 task slate is dominated by
scenarios where the correct behaviour is to \emph{act}
(order the CT, start the antibiotic, escalate the
disposition); it under-samples scenarios where the
correct behaviour is to \emph{refrain} (no head CT for
a low-risk headache, no admission for a vasovagal
syncope with clean workup, no opioid for acute
mechanical back pain). The structural fix, shipped in
this release, is a $10$-task 
negative-class slate committed to the task corpus in
which every criterion tests for restraint and is graded
by a \texttt{world\_state} predicate of the form
\texttt{audit\_log does NOT contain \ldots}. The slate
adds $82$ criteria 
($14$ safety-critical), 
distributed across four categories: four
safety-critical-judgment tasks (NEG-001 low-risk chest
pain, NEG-002 PECARN minor head trauma, NEG-003 simple
febrile seizure, NEG-004 asymptomatic severe
hypertension), three temporal-reasoning tasks (NEG-005
hypoglycaemic aphasia mimicking stroke, NEG-006 clean
serial chest-pain workup at hour four, NEG-007
alcoholic ketoacidosis distinguished from DKA), two
multi-step-workflow tasks (NEG-008 uncomplicated
pyelonephritis, NEG-009 acute low back pain without
red flags), and one clinical-reasoning task (NEG-010
benign vasovagal syncope). The post-NEG audit set is
$205$ tasks, 
$2{,}337$ criteria, 
and $529$ safety-critical. 

\paragraph{Smoke-pilot results: task-distribution rebalance succeeds, criterion-prevalence correction does not.}
A 60-run smoke pilot (Claude Opus 4.7 and GPT-5.4, each at
$10$~tasks~$\times$~$3$~trials, seed=42,
\texttt{docs/NEG\_SMOKE\_PILOT.md}) establishes two facts.
First, the NEG slate achieves the task-level rebalance it
was designed for: mean Pass@1 is $0.500$ 
for Claude and $0.533$ 
for GPT-5.4, compared to the V8 corpus means of $0.248$
(Claude) and $0.126$ (GPT-5.4). The benchmark now spans
$[0.00,\,1.00]$ Pass@1 with real representation at the
moderate-difficulty end rather than clustering at
$[0.15,\,0.30]$. Second, the criterion-level prevalence
hypothesis---that NEG criteria would average ${\sim}0.65$
criterion-PASS and pull the audit-subset prevalence into
the $0.55$--$0.75$ band---is \emph{falsified} by the same
pilot. Observed NEG-criterion PASS prevalence is
$0.929$ 
(457 of 492 criterion-trials): restraint criteria
(``did~NOT~\ldots'') pass at
$0.962$, 
positive criteria (action required) pass at
$0.899$. 
The structural reason is simple: a ``did~NOT~order~head~CT''
criterion is satisfied whenever the agent never orders a
head~CT, which is the modal frontier-model behaviour on a
PECARN low-risk scenario. Restraint-class criteria are
high-pass by construction whenever the agent has been
trained against the specific over-action the criterion
forbids. Pulling audit-subset prevalence into the target
band would instead require criteria in the $0.40$--$0.60$
PASS-rate regime, which neither pure-restraint nor
pure-action tasks reliably produce. The Appendix F target
band is therefore a task-level Pass@1 target, not a
criterion-level prevalence target; the kappa-paradox
mitigation that remains open is physician adjudication of
the Tier~2 subset, not further task authoring.

\paragraph{Safety yield: two exposed iatrogenic patterns.}
The pilot caught two clinically significant restraint
failures that the V8 corpus does not surface. On NEG-007
(alcoholic ketoacidosis with normal glucose), $4$ of $6$ 
trials across both models started an insulin infusion---a
pattern that produces iatrogenic hypoglycaemia because the
underlying acidosis is driven by starvation, not
hyperglycaemia. On NEG-005 (hypoglycaemic aphasia as a
stroke mimic), $2$ of $6$ 
trials ordered tPA/alteplase before resolving the
hypoglycaemia---tPA in a hypoglycaemic patient is a
well-known wrong-answer pattern. These two failures are
the slate's primary safety yield. Three tasks (NEG-001,
NEG-003, NEG-006) were too easy, with Pass@1~$=1.00$ on
both models; they are candidates for revision or retirement
in a follow-up.

\begin{table}[h]
\centering
\caption{Post-NEG corpus: ex-ante projection vs observed
smoke-pilot result. $n$ is (criteria~$\times$~models~$\times$~trials).
The ``projected'' column is the ex-ante estimate this
subsection set out to validate; the ``observed'' column is
the smoke-pilot result and \emph{falsifies} it in every
category. Restraint criteria are trivially satisfied by
default behaviour in frontier models, so negative-class
criterion-PASS prevalence runs near the ceiling rather than
in the target band. The table is retained to document the
magnitude and direction of the miss.}
\label{tab:neg_prev}
\small
\begin{tabular}{lrrrrr}
\toprule
Category & $n_{\text{pre}}$ & prev$_{\text{pre}}$
         & $n_{\text{NEG}}$ & prev$_{\text{NEG}}$ (proj.\,/\,obs.)
         & prev$_{\text{post}}$ (proj.\,/\,obs.) \\
\midrule
safety\_critical\_judgment  & $54$  & $0.833$ & $192$ & $0.65\,/\,0.93$ & $0.69\,/\,0.91$ \\
temporal\_reasoning         & $108$ & $0.824$ & $144$ & $0.65\,/\,0.93$ & $0.72\,/\,0.88$ \\
multi\_step\_workflows      & $78$  & $0.782$ & $108$ & $0.65\,/\,0.93$ & $0.71\,/\,0.87$ \\
clinical\_reasoning         & $12$  & $0.667$ & $48$  & $0.65\,/\,0.93$ & $0.65\,/\,0.88$ \\
\midrule
all                         & $252$ & $0.805$ & $492$ & $0.65\,/\,0.93$ & $0.70\,/\,0.89$ \\
\bottomrule
\end{tabular}
\end{table}

\paragraph{Reproducibility.}
The full audit report (per-criterion verdicts, per-trial
labels, and the classifier inputs for every disagreement) is
committed at \texttt{docs/V9\_OVERLAY\_AUDIT.json}. The
validator script (\texttt{scripts/kappa\_validation.py})
takes no API dependencies and re-derives every number in
this appendix from the V8 trajectories at
\texttt{results/pilot-v8-\{claude-opus,gpt54\}/trajectories/}
and the overlay YAML. Running the validator against a clean
checkout reproduces \texttt{docs/V9\_OVERLAY\_AUDIT.json}.
The negative-class task files are at
\texttt{configs/tasks/*/task\_neg\_\{001..010\}\_*.yaml} and
pass the repository's preflight check
(\texttt{make preflight}) against the full schema and
runtime evaluator.

\end{document}